\pdfoutput=1
\documentclass[11pt]{article}

\usepackage[final]{acl}

\usepackage{times}
\usepackage{latexsym}
\usepackage{graphicx}
\usepackage{subcaption}
\usepackage{color}
\usepackage{amsmath}
\usepackage{booktabs}
\usepackage{multirow}
\usepackage{bbding}

\usepackage[T1]{fontenc}
\usepackage[utf8]{inputenc}

\usepackage{microtype}

\usepackage{inconsolata}

\usepackage{graphicx}

\title{HELPD: Mitigating Hallucination of LVLMs by Hierarchical Feedback Learning with Vision-enhanced Penalty Decoding}

\author{Fan Yuan\textsuperscript{\rm 1,2}, Chi Qin\textsuperscript{\rm 1,2}, Xiaogang Xu\textsuperscript{\rm 3}, Piji Li\textsuperscript{\rm 1,2}$^{\ast}$ \\
\textsuperscript{\rm 1} College of  Artificial Intelligence, \\
Nanjing University of Aeronautics and Astronautics, Nanjing, China \\
\textsuperscript{\rm 2} MIIT Key Laboratory of Pattern Analysis and Machine Intelligence, Nanjing, China \\
\textsuperscript{\rm 3} The Chinese University of Hong Kong, Hong Kong, China\\
\texttt{\{fanyuan, qinchi, pjli\}@nuaa.edu.cn}, 
\texttt{xiaogangxu00@gmail.com}}

\begin{document}
\maketitle
\renewcommand{\thefootnote}{\fnsymbol{footnote}}
\footnotetext[1]{Corresponding author.}
\renewcommand{\thefootnote}{\arabic{footnote}}
\begin{abstract}
Large Vision-Language Models (LVLMs) have shown remarkable performance on many visual-language tasks. However, these models still suffer from \emph{multimodal hallucination}, which means the generation of objects or content that violates the images. Many existing work detects hallucination by directly judging whether an object exists in an image, overlooking the association between the object and semantics. To address this issue, we propose Hierarchical Feedback Learning with Vision-enhanced Penalty Decoding (HELPD). This framework incorporates hallucination feedback at both object and sentence semantic levels. Remarkably, even with a marginal degree of training, this approach can alleviate over 15\% of hallucination. Simultaneously, HELPD penalizes the output logits according to the image attention window to avoid being overly affected by generated text. HELPD can be seamlessly integrated with any LVLMs. Our experiments demonstrate that the proposed framework yields favorable results across multiple hallucination benchmarks. It effectively mitigates hallucination for different LVLMs and concurrently improves their text generation quality. \footnote{Code is available at \url{https://github.com/F-Yuan303/HELPD}}

% without impacting sentence length and concurrently improves their text generation quality.
\end{abstract}

%===========================================   Introduction   ==============================================================================

\section{Introduction}
\label{introduction}
Large Language Models (LLMs) (\citealp{brown2020, achiam2023gpt, llama1, llama2}), guided by human instruction, have demonstrated impressive performance in numerous Natural Language Processing (NLP) tasks (\citealp{qin2023chatgpt}). In light of the success of LLMs, researchers aspire to integrate the powerful capabilities of LLMs into multimodal domains, consequently introducing Large Vision-Language Models (LVLMs) (\citealp{blip2, instructblip, minigpt4, mplugowl2, llava}). Particularly, GPT-4 (\citealp{achiam2023gpt}) has been endowed with the capability to engage in complex, image-based dialogues with humans, while also being proficient in resolving a series of visual-language tasks.

\begin{figure}[t]
    \centering
    \includegraphics[width=0.95\columnwidth]{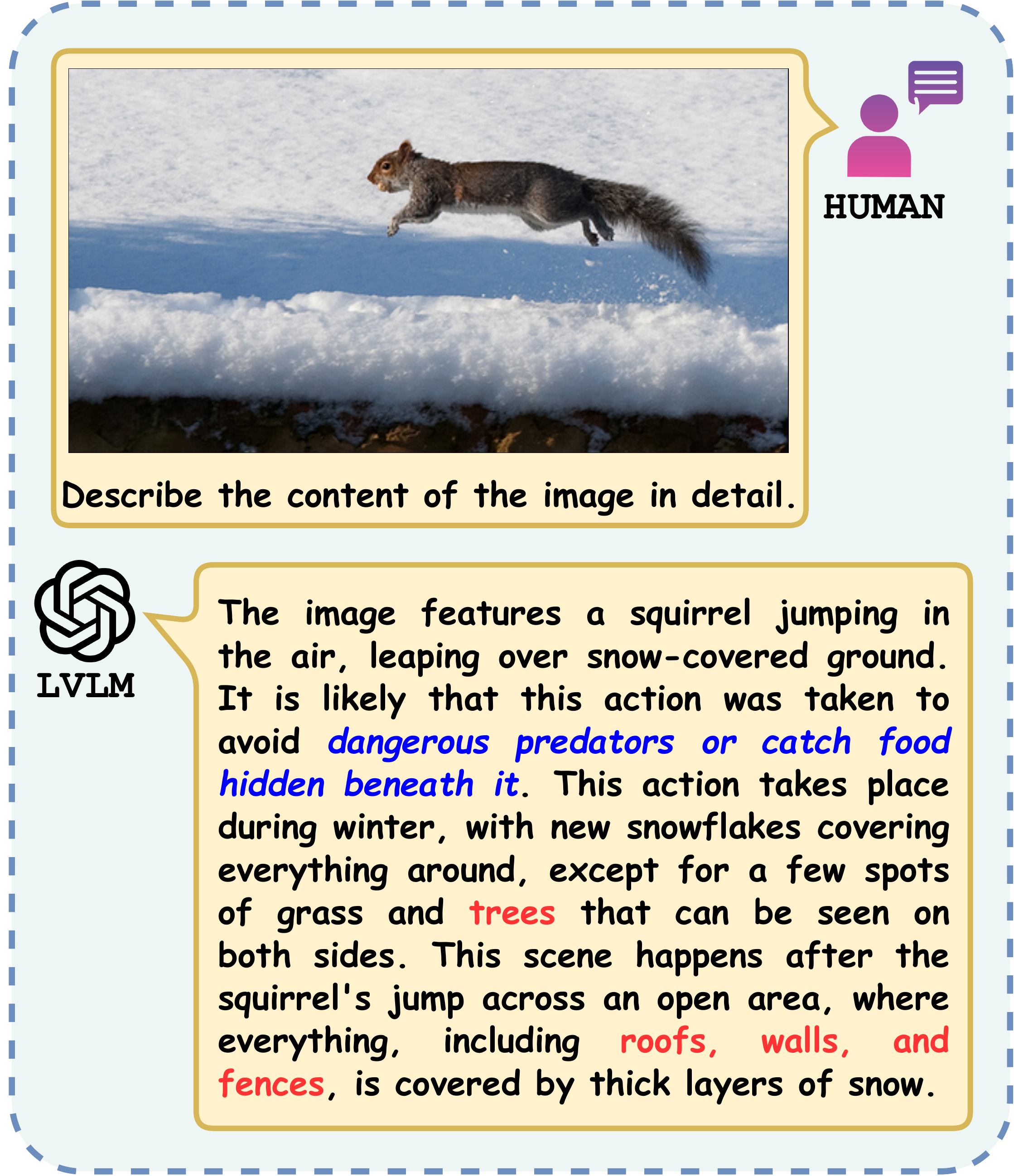}
    \caption{\textbf{A case of LVLM hallucination.} The parts marked in red are, in fact, hallucinations. The parts marked in blue would be mistaken for hallucinations by detection methods that focus only on objects.}
    \label{instruction_case}
\end{figure}

Despite the fact that LVLMs have achieved quite considerable results on various tasks, problems with these models have gradually emerged. Within these problems, the hallucination (\citealp{rohrbach2019object, pope}) has attracted significant attention. This is a phenomenon that LVLMs tend to generate content contradictory to the image, such as non-existent objects. In order to alleviate this phenomenon, many explorations have been carried out in recent work (\citealp{pope, wang2023evaluation, liu2023mitigating, zhou2023analyzing, zhai2023halle, lee2023volcano, huang2023opera, wang2023mitigating}). CoVe (\citealp{dhuliawala2023chain}) proposes a Chain-of-Verification method, it first generates verification questions, then executes them to check for hallucination, and finally gets a revised response. Additionally, some approaches aim to alleviate the hallucination from the decoding strategy (\citealp{o2023contrastive, chuang2023dola, huang2023opera}). Dola decoding (\citealp{chuang2023dola}) is a strategy of contrasting the mature layer and the immature layer of the model, followed by the determination of the next token based on the differences in logits. Opera (\citealp{huang2023opera}) is a recently proposed decoding method that employs an Over-trust Penalty to determine the occurrence of hallucination, and utilizes a Retrospection-Allocation rollback mechanism for decoding.

\begin{figure}[t]
    \centering
    \includegraphics[width=\columnwidth]{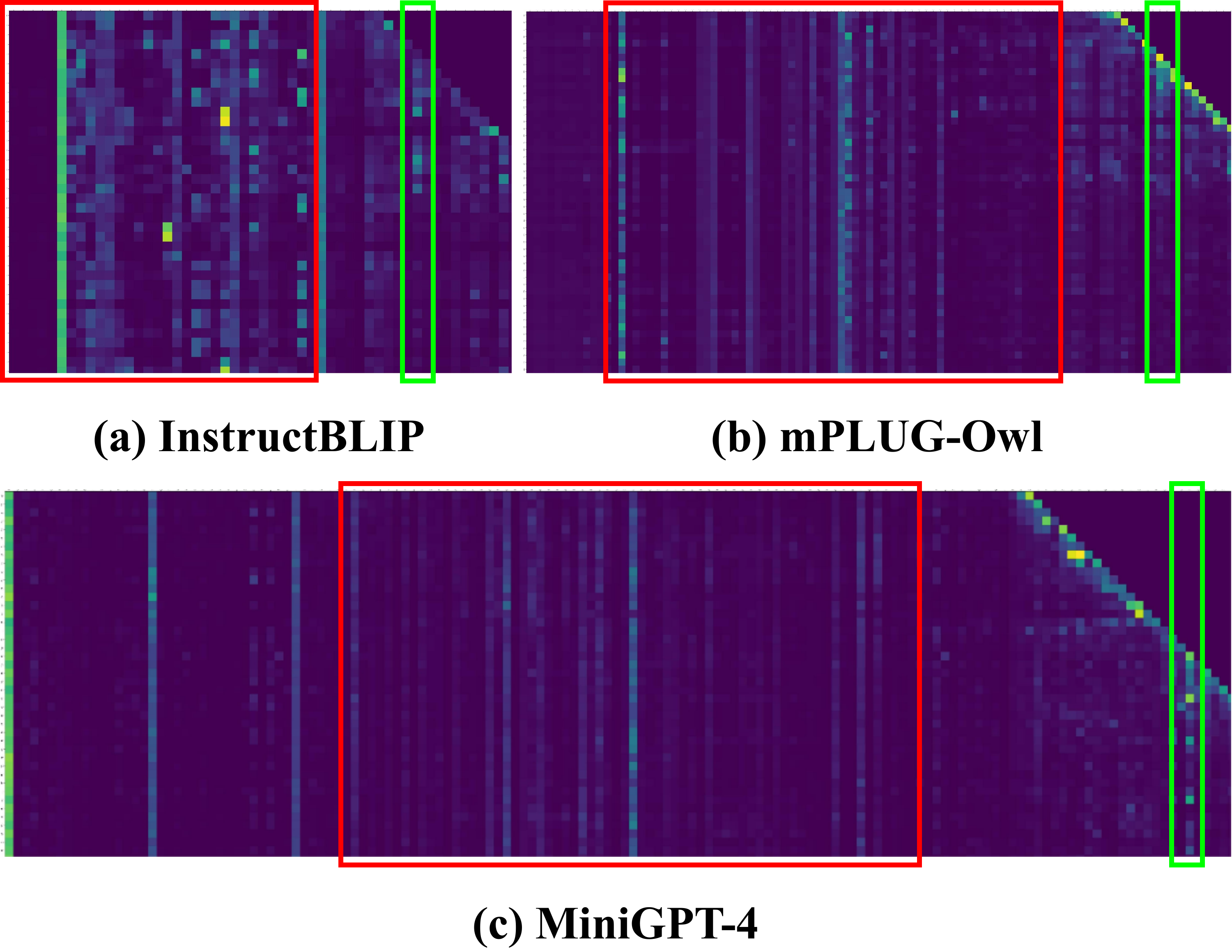}
    \vspace{-20pt}
    \caption{\textbf{Attention visualization of LVLMs.} For the same input, each image represents the attention matrix of a specific LVLM generation instance. \emph{Red} indicates the attention of the image, while \emph{green} represents the phenomenon of ``Over-trust" in the generated text. 
    }
    \vspace{-20pt}
    \label{vision_attention}
\end{figure}

However, most of the existing work focuses on alleviating object-level hallucinations. During this process, some methods excessively concentrate on whether the generated objects exist in the image, neglecting the association between these objects and the semantics of the whole sentence. As illustrated in Figure~\ref{instruction_case}, words marked in red, such as ``trees'', are real object hallucinations. Solely considering the presence of objects, the parts marked in blue, such as ``predators'' and ``food'', would also be defined as hallucinations. Nevertheless, combined with context and semantics, such a definition is deemed inappropriate. Meanwhile, as indicated by the green box in Figure~\ref{vision_attention}, ``Over-trust'' (\citealp{huang2023opera}) does exist in LVLMs, which means certain generated tokens receive excessive attention, leading to a subsequent generation that deviates from the image. We initially assumed that insufficient focus on the visual part of the input might be one cause of this phenomenon. However, further observation of the attention matrix of these models reveals strong focus on the visual input (see red boxes in Figure~\ref{vision_attention}). This indicates that considering the over-trust penalty only accounts for the impact of the text, additional focus on the image is therefore required to balance it.

% This indicates that during the decoding process, the model needs to focus more on the image.

Based on the observations above, we propose HELPD, a novel LVLM framework that utilizes \textbf{H}ierarchical Fe\textbf{E}dback \textbf{L}earning with Vision-enhanced \textbf{P}enalty \textbf{D}ecoding. We note the necessity of integrating both the inherent properties of an object and semantic meaning to determine the presence of hallucination. Thus, we propose the hierarchical feedback learning, which only requires a small amount of training, and we add this feedback mechanism at the end of the training period. On the one hand, the collection of objects is extracted from the sampled sentences and label sentences, and the object-level feedback is obtained through the comparison of the object sets. On the other hand, leveraging the powerful few-shot inference capabilities of GPT-4, we conduct a semantic comparison to obtain sentence-level feedback.

The manner of sampling constitutes a crucial component in the hierarchical feedback learning process. Opera decoding (\citealp{huang2023opera}) predicts the next word by subtracting the over-trust penalty score from the logits, where the penalty score is computed based on the attention window of the generated text, disregarding the potent influence of visual attention. Consequently, we propose the Vision-Enhanced Penalty Decoding, which incorporates visual attention into the penalty score computation and makes the final logits place more emphasis on the image input. This approach effectively mitigates an over-reliance on the textual modality during the decoding process, and enhances the influence of visual modality, thereby alleviating the hallucination.

% With the experiments on various benchmarks and metrics, the HELPD has been shown to effectively mitigate the occurrence of hallucination across different LVLMs. 

Our contributions can be summarized as follows:
\begin{itemize}
  \item We propose a hierarchical feedback learning method, incorporating object-level and sentence-level hallucination feedback. It can mitigate the occurrence of hallucinations with only a minimal amount of training.
  \item With the analysis of the attention matrix during decoding, we introduce the vision-enhanced penalty decoding to enhance the influence of images on the generation process.
  \item Extensive experimental results indicate that our proposed framework shows better performance on multiple hallucination metrics and can effectively alleviate the hallucination of LVLMs.
\end{itemize}

%===========================================   Related work   ==============================================================================

\section{Related Work}
\subsection{Large Vision-Language Models (LVLMs)}
Owing to the success of Large Language Models (LLMs) (\citealp{brown2020, achiam2023gpt, llama1, llama2, chung2022scaling, zeng2022glm, sun2021ernie, yang2023baichuan}) in many Natural Language Processing (NLP) tasks (\citealp{qin2023chatgpt}), many researchers have endowed it with multimodal perception capabilities. Among these, Large Vision-Language Models (LVLMs) (\citealp{blip2, instructblip, minigpt4, mplugowl2, llava, achiam2023gpt, peng2023kosmos, team2023gemini}) have shown particularly notable performance.

LVLMs primarily consist of three components: a visual encoder, a modality alignment module, and a Large Language Model (LLM) (\citealp{yin2023survey, zhang2024mm}). Visual encoders include Vision Transformers (ViT) (\citealp{dosovitskiy2020image}), CLIP ViT (\citealp{radford2021learning}), and others (\citealp{brock2021high, fang2023eva}). Specifically, ViT splits images into patches, which are then input into Transformer blocks through linear mapping for feature learning. Since there exits a modality gap between the visual encoders and the LLMs, the modal alignment module is required as a bridge. Models such as Flamingo (\citealp{alayrac2022flamingo}), BLIP-2 (\citealp{blip2}), and InstructBLIP (\citealp{instructblip}) apply the Q-former, a method that extracts visual features in a query-based manner by employing a set of learnable vectors. Another more direct method involves using a linear interface for modality alignment. For instance, LLaVA (\citealp{llava}) employs a linear layer to map images to the textual embedding space.

% Typically, the training of LVLMs is divided into two stages: multimodal pretraining and multimodal instruction fine-tuning. The first stage utilizes large-scale image-text datasets for for modal alignment. The second stage comprises Supervised Fine-Tuning (SFT) and Reinforcement Learning from Human Feedback (RLHF). This stage is designed to align with human intentions, and to enhance the interactive capabilities of LVLMs.

\subsection{Hallucination in LVLMs}
Multimodal hallucination is a significant challenge faced by LVLMs, severely impairing the reliability and robustness of these models. It typically manifests as generating content that is inconsistent with the image or contradicts common sense. Generally, hallucinations can be divided into Intrinsic Hallucination and Extrinsic Hallucination. Intrinsic hallucination refers to the generation of content that conflicts with the input. On the other hand, extrinsic hallucination represents the generation of additional content that does not actually exist, such as objects not present in the image.

Recently, numerous efforts have been dedicated to the elimination of multimodal hallucination  (\citealp{pope, wang2023evaluation, liu2023mitigating, zhou2023analyzing, zhai2023halle, lee2023volcano, huang2023opera, wang2023mitigating, DBLP:conf/aaai/GunjalYB24, DBLP:journals/corr/abs-2311-16839}). CHAIR (\citealp{rohrbach2019object}) is an early proposed metric for evaluating object hallucinations in image captioning tasks. It assesses the degree of hallucination by calculating the proportion of objects that appear in the generated descriptions but not in the image itself. POPE (\citealp{pope}) introduces a polling-based object probing evaluation method, which assesses the degree of hallucination based on the responses to questions like ``Is there a <object> in the image?'' that are posed based on the objects. CoVe (\citealp{dhuliawala2023chain}) introduces a chain-of-verification that considers its own responses and self-corrects hallucination. \citet{liu2023mitigating} conducts visual instruction tuning on LVLMs with the newly proposed LRV-Instruction dataset to mitigate hallucinations.

% Dola  (\citealp{chuang2023dola}) corrects the hallucination in the decoding stage through layer contrastive decoding.

%===========================================   Method   ==============================================================================
\begin{figure*}[t]
\includegraphics[width=\linewidth]{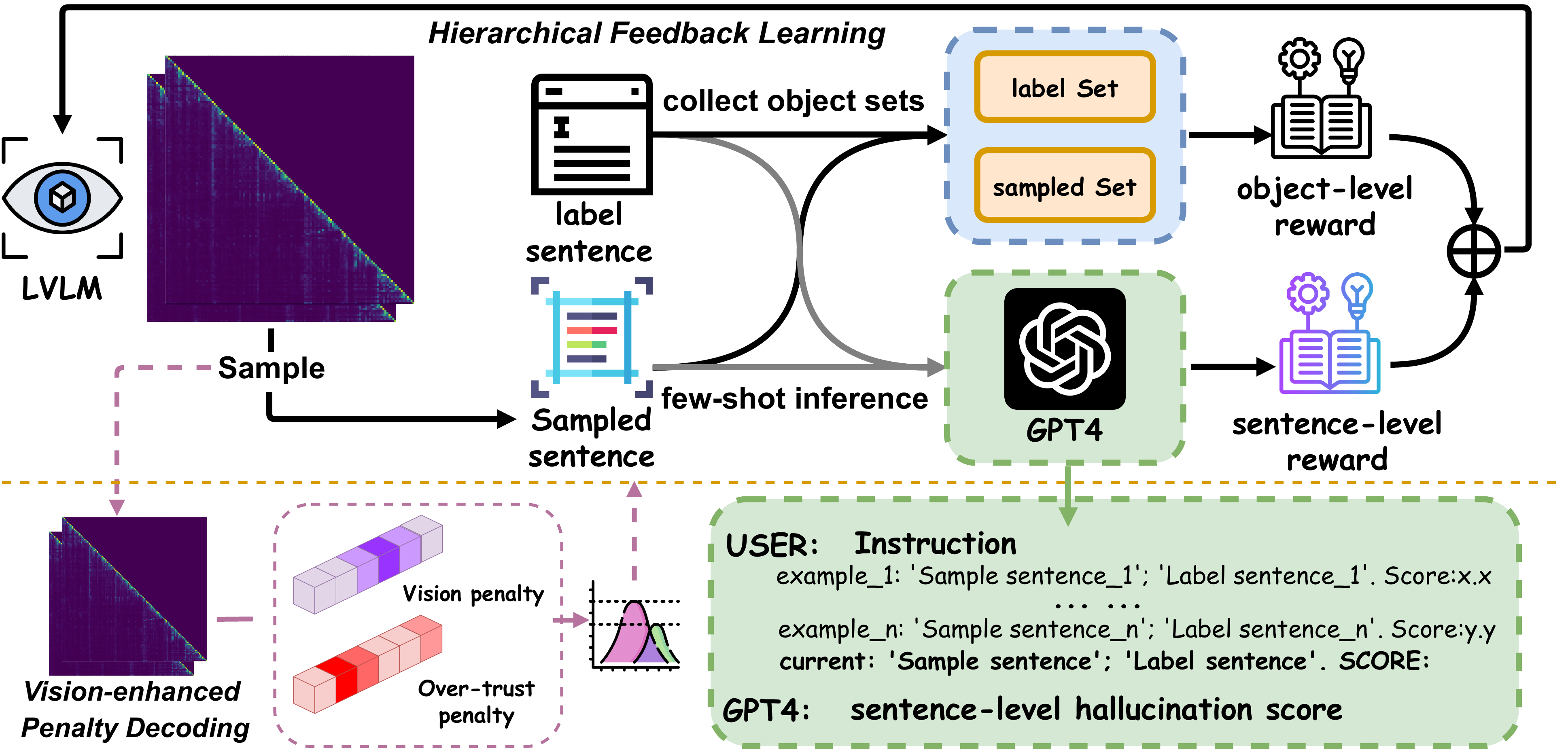 }
\caption{\textbf{This diagram illustrates the framework of HELPD.} The Hierarchical Feedback Learning detects hallucination by obtaining object-level feedback from comparing object sets extracted from sampled and label sentences, and sentence-level feedback through semantic comparison using GPT-4's few-shot inference capabilities. To improve the effectiveness of sampling, the Vision Penalty Decoding augments the over-trust penalty score with a vision-enhanced penalty score, making the final logits closer to the image.} 
\vspace{-5pt}
\label{method}
\end{figure*}

\section{Method}
\subsection{Hierarchical Feedback Learning}
\label{Hierarchical_Feedback_Learning}
As we have illustrated in Section~\ref{introduction}, to determine the occurrence of hallucination, it is necessary to consider not only whether the mentioned object appears in the image, but also to judge whether it is a reasonable association in combination with semantics.
To address the aforementioned issue, we propose Hierarchical Feedback Learning, a learning method that enhances the model's intrinsic ability to avoid hallucination through different granularity hallucination detection feedback (see Figure~\ref{method}).

In practice, we conduct minimal further training on the LVLMs and incorporate this feedback-learning mechanism towards the end of the training process. More specifically, after every fixed number of training steps, we sample the logits of the model's output to obtain the actions (which means the sampled tokens) $A = \{a_{ij}\}_{j=1}^{t},\quad i=1,\ldots,b$, where $t$ is the length of the sampled sentence and $b$ is batch size. With the help of NLTK \footnote{https://www.nltk.org/install.html} and GPT-4, we extract objects from the sampled sentence and label sentence, respectively, obtaining the sampled object set $S_{sam}=\{obj_1, obj_2, \dots, obj_m\}$ and the label object set $S_{lab}=\{obj_1, obj_2, \dots, obj_n\}$, where $m$ and $n$ represent the number of objects. Subsequently, we calculate the F1 score of these two sets as the object-level feedback scores $R_{obj}$:
\vspace{-5pt}
\begin{equation}
    \text{Precision}=\frac{|S_{sam} \cap S_{lab}|}{|S_{sam} \cap S_{lab}| + |S_{sam} \setminus S_{lab}|}
\end{equation}
\begin{equation}
    \text{Recall} = \frac{|S_{sam} \cap S_{lab}|}{|S_{sam} \cap S_{lab}| + |S_{lab} \setminus S_{sam}|}
\end{equation}
\begin{equation}
    R_{obj} = 2 * \frac{\text{Precision} * \text{Recall}}{\text{Precision} + \text{Recall}}
\end{equation}
Sentence-level feedback is obtained through the few-shot inference capability of GPT-4. We provide a detailed evaluation method and pre-annotate several sentence pairs as context (see Appendix~\ref{Appendix:prompt} for detail) to instruct it on discerning hallucination from semantics. The score ranging from 0 to 1, returned by GPT-4, is defined as $R_{sen}$. 

Given that $R_{sen}$ and $R_{obj}$ are non-differentiable, they cannot be directly incorporated into training using the gradient method. Inspired by (\citealp{SuttonMSM99}), we introduce the reinforce algorithm to handle this problem. Specifically, based on the tokens sampled, we first retrieve their corresponding log probabilities from the original logits: 
\begin{equation}
    P_{i,j} = \log\left(\frac{e^{logits_{i,j,A_{i,j}}}}{\sum_{k=1}^{V} e^{logits_{i,j,k}}}\right),
\end{equation} 
where $i$ is the index within the batch, $j$ is the index within the sequence length, $A_{i,j}$ is the corresponding sampled action, and $k$ is the index within the vocabulary of size $V$. A hyperparameter $\sigma$ is set to determine the relative importance of the two types of feedback mentioned above (see Equation \eqref{sigma}). By summing the product of the feedback and the corresponding log probabilities of the actions, we obtain a negative weighted log-likelihood loss. To prevent the loss from infinitely increasing with the number of actions, the total loss is divided by the number of sampled actions to yield the loss function for reinforce algorithm, denoted as $\mathcal{L}_{\text{RL}}$:
% Finally, the total loss is divided by the total number of sampled actions to ensure that the loss does not infinitely increase with the number of actions.
\vspace{-5pt}
\begin{equation}
\label{sigma}
    R_i = \sigma R_{sen,i} + (1-\sigma) R_{obj,i},
\end{equation}
\begin{equation}
\mathcal{L}_{\text{RL}} = -\frac{1}{N}\sum_{i=1}^{b} \sum_{j=1}^{t} P_{ij} \cdot R_i.
\end{equation}
In the early stage of training, we employ cross-entropy loss $\mathcal{L}_{\text{CE}}$. When the training step reaches $c * \text{total steps}$, $\mathcal{L}_{\text{RL}}$ is added to the total loss. The total loss is defined as:
\vspace{-8pt}
\begin{equation}
\mathcal{L}_{\text{CE}}=-\sum\limits_{h}^{H} \log P\left(x_h \mid x_{<h}\right),
\end{equation}
\vspace{-10pt}
\begin{equation}
\mathcal{L} = 
\begin{cases} 
\mathcal{L}_{\text{CE}}, & \text{if } \text{steps} < c * \text{total steps}, \\
\frac{\mathcal{L}_{\text{CE}}}{\|\mathcal{L}_{\text{CE}}\|} + \frac{\mathcal{L}_{\text{RL}}}{\|\mathcal{L}_{\text{RL}}\|}, & \text{otherwise},
\end{cases}
\end{equation}
where $x$ is the generated token.

\begin{figure}[t]
    \centering
    \includegraphics[width=\columnwidth]{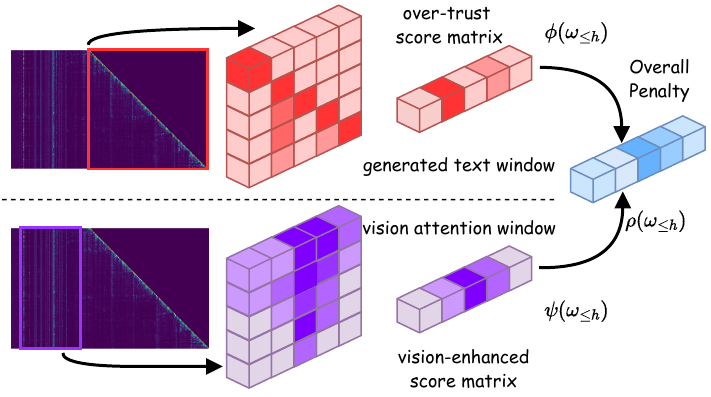}
    \vspace{-15pt}
    \caption{\textbf{The illustration of Vision-enhanced Penalty Decoding.} The total penalty is composed of the vision penalty and the over-trust penalty. The over-trust penalty is computed based on the generated text (the upper region), while the vision penalty is computed from the vision attention window (the lower area).}
    \vspace{-10pt}
    \label{decoding}
\end{figure}

\subsection{Vision-enhanced Penalty Decoding}
With the hierarchical feedback learning, we can effectively detect object-level and sentence-level hallucinations and correct them through back propagation. To obtain the sampled sentences, we need to sample each token from the model's logits. Based on our analysis of the attention matrix, we propose the Vision-enhanced Penalty Decoding based on Opera (\citealp{huang2023opera}).
\vspace{-5pt}
\paragraph{Over-Trust Logit Penalty.} First, we provide a brief description of the original process. It sets a local window of length $h$ for the attention matrix, where $h$ represents the length of the current generated sequence. Upon obtaining such a lower triangular matrix, it pads the upper triangular part with zeros and scales up the values to avoid excessively small values. Subsequently, it conducts column-wise multiplication on this matrix and select the maximum value of the vector as the over-trust penalty $\phi (\omega_{\le h})$. Finally, it subtracts this penalty from the original logits to predict the next token.

% , and incorporates this penalty with original logits to predict the next token.
\vspace{-5pt}
\paragraph{Vision-enhanced Penalty Decoding.} As illustrated in Section~\ref{introduction}, we note that the over-trust penalty establishes a local window whose size is limited to the length of the generated text. This approach can effectively extract over-trust patterns from past tokens, but it inadvertently amplifies the model's reliance on the text modality, thereby passively diminishing its focus on images. 

To foster a greater focus on images during the sampling process, we set an additional local window $\mathbf{W}_{l}^{h}$ beyond the local window of the over-trust penalty, as shown in Figure~\ref{decoding}, purposed for storing the image components within the attention matrix:
\begin{equation}
\label{attention_matrix}
   \mathbf{W}_{l}^{h}=\{\mathbf{w}^i\}_{i=1}^{h},\quad\text{s.t. }\mathbf{w}^i=\{\omega_{i,j}\}_{j=1}^{l},
\end{equation}
where $h$ is the length of the over-trust penalty window, $l$ means the length of the visual input within the attention matrix, and $\omega_{i,j}$ represents the attention weight from $i_{th}$ token to $j_{th}$ token. Subsequently, we conduct the column-wise multiplication on the $\mathbf{W}_{l}^{h}$ to obtain a vector of column-wise scores, which represents the accumulated attention values of image:
\vspace{-5pt}
\begin{equation}
    \psi(\omega_{\le h}) = \sum_{i=1}^{h} \omega_{i},\quad\text{s.t. }\omega_{i}=\prod_{j=1}^{l} \omega_{i,j},
\end{equation}
% \vspace{-5pt}
where $\psi(\omega_{\le h})$ means the vision-enhanced penalty.

% the final layer of attention weight $\mathbf{w}^{h}=\{\omega_{h,j}\}_{j=1}^{m}$ pertaining to the image component can be obtained, which is used for the generation of the next token.

% Then we define $\psi(\omega_{\le h}) = \sum_{j=1}^{m} \omega_{h,j}$ as the vision-enhanced penalty.

% \mathbf{w}^{h}

Given the difference in numerical magnitudes, the initial step involves scaling $\psi(\omega_{\le h})$ to match the order of magnitude of $\phi (\omega_{\le h})$, then calculating the overall penalty weight $\rho(\omega_{\le h})$, as: 
\vspace{-5pt}
\begin{equation}
\begin{aligned}
   \rho(\omega_{\le h})=\phi(\omega_{\le h})-\beta\psi(\omega_{\le h}), \\
   \quad\text{s.t. }\beta=\frac{\overline {\sum\limits_{j\le h} \phi(\omega_j)}}{\overline {\sum\limits_{j\le h} \psi(\omega_j)}}.
\end{aligned}
\end{equation}
Then, this penalty weight is added to the original logits for the prediction of the next token $\hat{x}_{h}$, as:
\vspace{-5pt}
\begin{equation}
 \hat{x}_{h} = \arg\max_{x \in V} [p(x|x_{<h}) -  \rho(\omega_{\le h})],
\end{equation}
where $V$ is the size of vocabulary and $x$ represents the predicted token.
%===========================================   Experiments   ==============================================================================
\begin{table*}[htbp] 
\centering
\scalebox{0.97}{
\begin{tabular}{ll|ccc|c|c}
\toprule
\textbf{POPE} &\textbf{Model} & Accuracy & Precision & Recall & F1 Score & Yes (\%)\\
\midrule

\multirow{6}{*}{\textit{Random}} 
& MiniGPT-4                 & 53.06 & 51.58 & 99.60 & 67.97 & 96.53  \\
& InstructBLIP              & 86.28 & 84.11 & 95.02 & 89.23 & 55.63 \\
& mPLUG-Owl2                & 87.71 & 89.13 & \textbf{85.92} & 87.50 & 48.21 \\  
& mPLUG-Owl2 (w/ ours)       & \textbf{88.02} & \textbf{89.90} & 85.67 & \textbf{87.73} & 47.62 \\
& LLaVA-1.5                 & 89.44 &	87.21 &	90.11 &	88.63 &	53.09 \\ 
& LLaVA-1.5 (w/ ours)        & \textbf{89.65} & \textbf{87.84} & \textbf{91.98} & \textbf{89.86} & 52.41 \\

\midrule
\multirow{6}{*}{\textit{Popular}} 
& MiniGPT-4                 & 50.53 &  50.26  &  99.60  & 66.81 &  89.06 \\
& InstructBLIP              & 81.67 & 74.12 & 93.31 & 82.61 & 65.43    \\
& mPLUG-Owl2                & 84.10 &	82.81 &	85.23 &	84.00 &	51.90  \\
& mPLUG-Owl2 (w/ ours)       & \textbf{85.67} &	\textbf{85.81} &	\textbf{85.27} &	\textbf{85.53} &	49.66  \\
& LLaVA-1.5                 & 84.91 &	81.02 &	90.71 &	85.59 & 55.51  \\ 
& LLaVA-1.5 (w/ ours)        & \textbf{85.79} &	\textbf{81.85} &	\textbf{91.97} &	\textbf{86.62} &	56.26  \\

\midrule
\multirow{6}{*}{\textit{Adversarial}}   
& MiniGPT-4                 & 50.46   &  50.23  & 99.61   & 66.78  & 99.13  \\
& InstructBLIP              & 72.12 & 65.63 & 95.27 & 77.32 & 73.26  \\
& mPLUG-Owl2                & \textbf{81.70} & 79.21 & \textbf{85.93} & 82.43 & 54.23   \\
& mPLUG-Owl2 (w/ ours)       & 81.64 & \textbf{80.28} & 85.65 & \textbf{82.87} & 53.31   \\
& LLaVA-1.5                 & 77.61 & 71.72 & 92.55 & 80.81 & 63.83   \\ 
& LLaVA-1.5 (w/ ours)        & \textbf{78.15} & \textbf{72.04} & \textbf{92.91} & \textbf{81.15} & 63.88   \\

\bottomrule
\end{tabular}
}
\caption{\textbf{Results of LVLMs under three evaluation settings of POPE on the validation set of MSCOCO.} ``Yes'' denotes the proportion of answering ``Yes'' to the given question. ``w/ ours'' means the application of HELPD.}
\vspace{-5pt}
\label{tab:POPE}
\end{table*}

\section{Experimental Setups}
\subsection{Hallucination Benchmarks}
\paragraph{CHAIR.}
Caption Hallucination Assessment with Image Relevance (CHAIR) (\citealp{rohrbach2019object}) is an evaluation metric employed for assessing hallucination in image captioning, and it is often used to evaluate LVLMs. CHAIR obtains scores for the degree of hallucination by calculating what proportion of objects generated are actually in the image according to the ground truth sentences and object segmentations.  Specifically, it computes the hallucination at both instance level (defined as CHAIR$_i$) and sentence level (defined as CHAIR$_s$):
\begin{equation}
    \text{CHAIR}_s = \frac{|\{\text{hallucinated objects}\}|}{|\{\text{all mentioned objects}\}|}, 
\end{equation}
\begin{equation}
    \resizebox{0.85\linewidth}{!}{$\text{CHAIR}_i = \frac{|\{\text{captions w/ hallucinated objects}\}|}{|\{\text{all captions}\}|}$},
\end{equation}
where CHAIR$_s$ represents the proportion of hallucinated objects among all mentioned objects, and CHAIR$_i$ denotes the proportion of captions with hallucinated objects among all captions.
\vspace{-5pt}
\paragraph{POPE.}
POPE (\citealp{pope}) converts hallucination assessment into asking the model to answer a series of true or false questions about whether an object is present in the image. Specifically, given an image set and the object annotations contained in each image, POPE will construct a series of triples consisting of images, questions, and answers. It considers three polling strategies by sampling the objects randomly, from popular objects, and among those frequently co-occurring objects, respectively. Finally, POPE involves 3K questions for the captions of 500 images and uses the Accuracy, Precision, Recall, and F1 scores for evaluation.
\vspace{-5pt}
\paragraph{GAVIE.}
GPT4-Assisted Visual Instruction Evaluation (GAVIE) (\citealp{liu2023mitigating}) is an approach to measure the hallucination without the need for human-annotated ground-truth answers. GPT-4 takes the generated captions with bounding box coordinates as the image content and compares human instructions and model response. Then, ask GPT-4 to score the answers based on two criteria: (1) Accuracy: whether the response hallucinates with the image content. (2) Relevancy: whether the response directly follows the instruction. It is composed of 1k questions and uses accuracy and relevancy for evaluation.
\vspace{-5pt}
\paragraph{MMHal-Bench.}
MMHal-Bench (\citealp{sun2023aligning}) has a focus on penalizing hallucinations with 96 image-question pairs, ranging in 8 question categories and 12 object topics from OpenImages (\citealp{kuznetsova2020open}). It uses GPT-4 to compare the model’s response to the correct answer based on the given object information. If the score is below 3, it is considered to have hallucinations.

\subsection{Baselines}
We use 4 recently released LVLMs as baselines: (1) MiniGPT4 (\citealp{minigpt4}); (2) InstructBLIP (\citealp{instructblip}); (3) LLaVA-1.5 (\citealp{llava}); (4) mPLUG-Owl2 (\citealp{mplugowl2}). All models above have been tuned on their visual instruction data.

\subsection{Implementation Details}
We randomly select 5,000 images from the training sets of MSCOCO 2014 (\citealp{lin2014microsoft}) and Flickr30k (\citealp{plummer2015flickr30k}). Given that each image corresponds to multiple short captions, we prompt GPT-4 to synthesize a longer caption for each image based on these short captions (see Appendix~\ref{Appendix:prompt}). Then, we employ LoRA-tuning (\citealp{hu2021lora}) and deepspeed zero stage 3 to conduct minimal training on LLaVA-1.5-7b and mPLUG-Owl2-7b for 1 epoch. We use the AdamW (\citealp{loshchilov2017decoupled}) optimizer for optimization purposes. The learning rate and weight decay are set to 0.0001 and 0.1, respectively. During the training process, we initiate a warm-up ratio of 0.03, after which we apply the cosine schedule to decay the learning rate. We set $\sigma$ to 0.6. The values of $c$ for LLaVA-1.5 and mPLUG-Owl2 are set to 0.7 and 0.8. Each model requires approximately 4 hours to train with 2 NVIDIA 3090 24Gb GPUs.
% It takes approximately 5 hours for each LVLM to train on 2 NVIDIA 3090 24Gb GPUs. 
% All of the experiments are conducted on 2 NVIDIA 3090 24Gb GPUs.

%===========================================   Results   ==============================================================================
\begin{table}[t]
\centering
\scalebox{0.83}{
\begin{tabular}{ll|cc|c}
\toprule
\textbf{Method} & \textbf{Model} & C$_s$ $\downarrow$ & C$_i$ $\downarrow$  & Len\\
\midrule
\multirow{4}{*}{\textit{Beam$_5$}}
& mPLUG-Owl2  & 46.6 & 14.5 & \textbf{68.4} \\
& mPLUG-Owl2 (w/ ours) & \textbf{22.4} &	\textbf{8.4} & 67.1\\
& LLaVA-1.5 & 15.4 & 8.2 & \textbf{63.2} \\
& LLaVA-1.5 (w/ ours) &	\textbf{14.6} & \textbf{6.1} & 57.8\\
\midrule

\multirow{4}{*}{\textit{Opera}}
& mPLUG-Owl2  & 39.8 & 13.1  & \textbf{64.5} \\
& mPLUG-Owl2 (w/ ours) & \textbf{21.4} &	\textbf{8.2} & 53.3 \\
& LLaVA-1.5 & 14.1 & 6.1 & 58.9 \\ 
& LLaVA-1.5 (w/ ours) &	\textbf{13.7} & \textbf{5.0} & \textbf{59.9} \\
\midrule

\multirow{4}{*}{\textit{Vep}}
& mPLUG-Owl2  & 36.2 & 13.0 & \textbf{65.1} \\
& mPLUG-Owl2 (w/ ours) & \textbf{20.6} &	\textbf{7.8} & 54.0 \\
& LLaVA-1.5 & 11.0 & 6.2 & 59.7 \\
& LLaVA-1.5 (w/ ours) &	\textbf{9.6}  & \textbf{4.9} & \textbf{60.8} \\

\bottomrule
\end{tabular}
}
\vspace{-5pt}
\caption{\textbf{CHAIR hallucination evaluation results.} ``w/ ours'' means the use of hierarchical feedback learning, and ``\textit{Vep}'' is the vision-enhanced penalty decoding.}
\label{tab:chair}
\end{table}

\begin{table}[t]
\centering
\scalebox{0.9}{
\begin{tabular}{l|c|c}
\toprule
\textbf{Model} & Relevancy &	Accuracy
 \\
\midrule
MiniGPT-4               & 3.84 & 5.35  \\
InstructBLIP            & 6.27 & 5.83  \\
mPLUG-Owl2              & 8.29 & 5.68  \\
mPLUG-Owl2 (w/ ours)    &\textbf{8.88} & \textbf{6.12}   \\
LLaVA-1.5               & 7.56 & 5.49  \\
LLaVA-1.5 (w/ ours)     & \textbf{7.98} & \textbf{6.01}  \\
\bottomrule
\end{tabular}
}
\vspace{-5pt}
\caption{\textbf{Evaluation results on GAVIE.} The metric scores of Relevancy and Accuracy are from 0 to 10. ``w/ ours'' means the application of HELPD.}
\vspace{-10
pt}
\label{tab:Gavie}
\end{table}

\begin{figure}[t]
    \centering
    \includegraphics[width=\columnwidth]{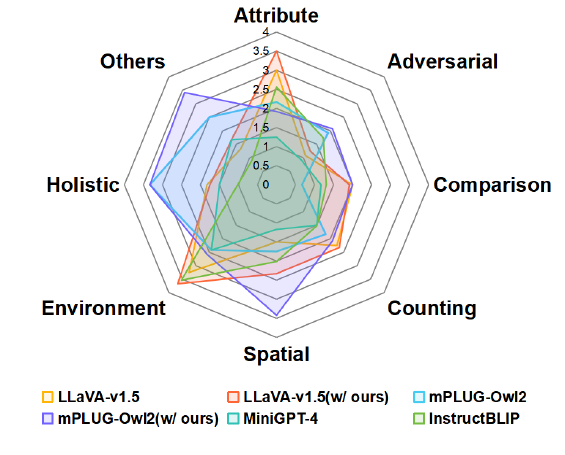}
    \vspace{-30pt}
    \caption{\textbf{Detailed performance of LVLMs on the eight categories in MMHAL-Bench,} where ``Overall'' indicates the averaged performance across all categories. ``w/ ours'' means the application of HELPD.}
    \vspace{-10pt}
    \label{MMhal}
\end{figure}

\section{Results}
\subsection{Main Results}
In general, it is noticeable that the application of HELPD with various LVLMs is able to enhance their performance across different evaluation metrics, compared to the original LVLMs.

% From the evaluation results on the POPE benchmark, it can be observed from Table~\ref{tab:POPE} that after hierarchical feedback learning, it shows improvements across the accuracy, precision, and F1 score of the LVLMs. 
Upon examining the results from the POPE benchmark, as detailed in Table~\ref{tab:POPE}, it is evident that the hierarchical feedback learning has led to enhancements in the accuracy, precision, and F1 score.This suggests that our proposed framework can provide effective hallucination detection and feedback during training, combining object entities and semantic information to guide the model in enhancing its ability to discern hallucinated objects. 

% Simultaneously, the trained model exhibits a declining trend in the ``Yes (\%)'', indicating that this feedback learning can also reduce LVLMs' preference for responding with 'yes'.

As shown in Table~\ref{tab:chair}, from the score of CHAIR$_s$ and  CHAIR$_i$, it is evident that with the help of HELPD, both mPLUG-Owl2 and LLaVA-1.5 have demonstrated varying degrees of hallucination reduction. Specifically, the trained mPLUG-Owl2, under various decoding methods, is able to reduce the CHAIR$_s$ by an average of 19.4 and the CHAIR$_i$ by an average of 5.4. This indicates that our proposed framework can effectively mitigate the generation of hallucinations, whether at the instance level or the sentence level. Moreover, it can be observed that the trained LVLMs do not exhibit significant fluctuations in the length of generated text. In most cases, LLaVA-1.5 can even increase the average generated length by 1.1 to 4.0. This illustrates that HELPD, while enhancing the anti-hallucination capabilities of LVLMs, does not excessively interfere with the generated length.

Table~\ref{tab:Gavie} shows the performance of different LVLMs on the GAVIE benchmark, which asks GPT-4 to pretend to be a smart teacher and scores (0-10) the answers according to the image content and instructions. The trained models achieve improvements in both accuracy and relevancy. Specifically, the trained mPLUG-Owl2 attains scores of 8.88 and 6.12 in relevancy and accuracy respectively, surpassing the provided baseline models. This demonstrates that our proposed framework can aid LVLMs in more directly following instructions, and the generated responses are more accurate concerning the image content.

Detailed performance of LVLMs on the eight categories in MMHAL-Bench is shown in Figure~\ref{MMhal}. It is evident that the trained models surpass their corresponding baseline models in performance across all eight question categories, and achieve a score of over 3 in five categories, including Object attribute and Spatial relation. This implies that their generated texts are somewhat informative and exhibit almost no hallucination.

\subsection{Further Analysis}
\label{Further_analysis}
\paragraph{Break-down Study of Hierarchical Feedback Learning.}
As illustrated in Section~\ref{Hierarchical_Feedback_Learning}. In order to detect hallucinations of different granularities during training and provide feedback for parameter updates, we introduce hallucination feedback at both the object and sentence levels. To verify whether these two types of feedback contribute to the mitigation of hallucination, we conduct ablation experiments, and the results are presented in Table~\ref{tab:feedback_ablation}. Both object-level and sentence-level feedback can aid in alleviating hallucination, making the generated text adhere more closely to instructions and rendering it more accurate concerning the image content. It can also be observed that, compared to object-level feedback, sentence-level feedback can more effectively enhance the model's ability to resist hallucination. We hypothesize that this is because object-level feedback is more uncontrollable, such as possible omissions in the process of object extraction, or score reductions due to the presence of synonyms. However, the sentence-level feedback generated by prompting GPT-4 can effectively compensate for the deficiencies of the object-level feedback, thereby enhancing the overall performance of hierarchical feedback learning.

\begin{table}[t]
\centering
\scalebox{0.8}{
\begin{tabular}{l|cc|cc|cc}
\toprule
\textbf{Model} & \textbf{$R_{obj}$} & \textbf{$R_{sen}$} & C$_s$ $\downarrow$ & C$_i$ $\downarrow$ & Rel & Acc\\
\midrule
\multirow{4}{*}{mPLUG-Owl2}
& \XSolidBrush  & \XSolidBrush  & 46.6  & 14.5  & 8.2  & 5.6    \\
& \Checkmark    & \XSolidBrush  & 31.1  & 11.2  & 8.4  & 5.9    \\
& \XSolidBrush  & \Checkmark    & 25.9  & 9.9   & 8.7  & 5.9    \\
& \Checkmark    & \Checkmark    & \textbf{22.4}  & \textbf{8.4}   & \textbf{8.8}  & \textbf{6.1}    \\
\midrule

\multirow{4}{*}{LLaVA-1.5}
& \XSolidBrush  & \XSolidBrush  & 15.4  & 8.2   & 7.5   & 5.4   \\
& \Checkmark    & \XSolidBrush  & 14.9  & 7.1   & 7.6   & 5.7   \\
& \XSolidBrush  & \Checkmark    & 15.8  & 6.8   & 7.7   & 5.8   \\ 
& \Checkmark    & \Checkmark    & \textbf{14.6}  & \textbf{6.1}   & \textbf{7.9}   & \textbf{6.0}   \\

\bottomrule
\end{tabular}
}
\vspace{-5pt}
\caption{\textbf{Ablation results on different levels of feedback on CHAIR and GAVIE.} $R_{obj}$ and $R_{sen}$ represent object-level and sentence-level feedback, respectively.}
\vspace{-10pt}
\label{tab:feedback_ablation}
\end{table}

\vspace{-5pt}
\paragraph{The Timing of Incorporating Hierarchical Feedback Learning.}
To investigate at which stage of training the integration of hierarchical feedback learning can better enhance the model's anti-hallucination capabilities, we also conduct an ablation study on the hyperparameter $c$. The experimental results of LLaVA-V1.5 on the random set of POPE are shown in Table~\ref{tab:timing_ablation}, with more details available in the Appendix~\ref{Appendix:results}. It indicates that LLaVA-V1.5 exhibits fewer hallucinations when $c=0.7$, while mPLUG-Owl2 performs better when $c=0.8$. Therefore, we default to assigning $c=0.7$ for LLaVA-V1.5 and $c=0.8$ for mPLUG-Owl2.
\vspace{-5pt}
\paragraph{Different Decoding Strategy.}
Based on the observations of the attention matrix, we propose the vision-enhanced penalty decoding based on opera. To validate its effectiveness, we conduct an ablation study on LVLMs. The experimental results are shown in Table~\ref{tab:chair} and Table~\ref{tab:decoding_ablation}. As can be observed, compared to the baseline decoding strategy, the vision-enhanced penalty decoding demonstrates superior performance on benchmarks such as POPE and CHAIR, and has a smaller impact on the length of the generated text. It should be noted that this decoding strategy pays more attention to the hallucination performance of long texts.

% , thus showing a smaller increase in binary evaluations such as POPE.

% default to assigning $c=0.7$ for LLaVA-V1.5 and $c=0.8$ for mPLUG-Owl2.

\begin{table}[t] 
\centering
\scalebox{0.92}{
\begin{tabular}{l|c|cc|c}
\toprule
\textbf{Model} &\textbf{$c$} & Precision & Recall & F1 Score \\
\midrule

% \multirow{5}{*}{mPLUG-Owl2}
% & 0.6   & 51.58 & 99.60 & 67.97   \\
% & 0.7   & 84.11 & 95.02 & 89.23  \\
% & 0.8   & 89.13 & \textbf{85.92} & 87.50  \\  
% & 0.9   & \textbf{89.90} & 85.67 & \textbf{87.73}  \\ 

% \midrule
\multirow{5}{*}{LLaVA-1.5} 
& 0.6   & 87.01 & 92.05 & 89.45  \\
& 0.7   & \textbf{87.84} & 91.98 & \textbf{89.86}  \\
& 0.8   & 86.21 & 93.07 & 89.50  \\  
& 0.9   & 86.09 & \textbf{93.33} & 89.56  \\
& 1.0   & 86.11 & 91.33 & 88.74 \\

\bottomrule
\end{tabular}
}
\vspace{-5pt}
\caption{\textbf{Ablation results on the timing of incorporating HELPD.} We only show the random set results on LLaVA-v1.5, more details can be seen in Appendix~\ref{Appendix:results}.}
\vspace{-5pt}
\label{tab:timing_ablation}
\end{table}

\begin{table}[t] 
\centering
\scalebox{0.65}{
\begin{tabular}{l|c|c|c|c}
\toprule
\textbf{Method} & MiniGPT-4 & InstructBLIP & mPLUG-Owl2 & LLaVA-v1.5 \\
\midrule

Nucleus     & 58.6   & 78.9 & 82.9 & 82.3   \\
Beam$_5$    & 69.2   & 82.1 & 84.7 & 84.7   \\
Opera       & 73.3   & 84.7 & 85.1 & 85.4    \\  
Vep         & \textbf{74.1}   & \textbf{85.0} & \textbf{85.3} & \textbf{85.6}   \\

\bottomrule
\end{tabular}
}
\vspace{-5pt}
\caption{\textbf{Ablation results on the decoding strategy.} We exhibit the average F1-score computed on random, popular, and adversarial splits of POPE.}
\vspace{-10pt}
\label{tab:decoding_ablation}
\end{table}

%===========================================   Conclusion   ==============================================================================
\section{Conclusion}
In this paper, we aim to alleviate hallucinations in Large Vision-Language Models (LVLMs), and propose the HELPD framework, which employs the hierarchical feedback learning for small amounts of training on the model. To enhance attention to the visual modality, we also propose a vision-enhanced penalty decoding strategy. To evaluate the effectiveness of our approach, we conduct evaluations on numerous benchmarks. Experimental results demonstrate that our proposed framework effectively mitigates hallucination for different LVLMs without impacting sentence length and concurrently improves their text generation quality. Future work could focus on a more comprehensive evaluation of hallucination at different granularities. 

% Given the differences and diversity among modalities, more attention should be paid to the alignment issue between modalities. This potentially serves as a more direct way to alleviate hallucination.

%==========================================================================================================================================
\section*{Limitations}
Although HELPD effectively mitigates the hallucination in VLVMs, it remains subject to certain limitations. Firstly, to further train LVLMs, even for minimal training, a rich corpus of modality-aligned data is required. Secondly, compared to traditional decoding strategies, our proposed vision-enhanced penalty decoding may slightly increase decoding time, thereby potentially limiting inference speed.

\section*{Ethics Statement}
The data (\citealp{lin2014microsoft}; \citealp{plummer2015flickr30k}) used in our work is all drawn from open-source datasets. The data and text involved in our research do not involve private information and social issues.

\section*{Acknowledgments}
This research is supported by the National Natural Science Foundation of China (No.62476127, No.62106105),  the Natural Science Foundation of Jiangsu Province (No.BK20242039), the CCF-Baidu Open Fund (No.CCF-Baidu202307), the CCF-Zhipu AI Large Model Fund (No.CCF-Zhipu202315), the Fundamental Research Funds for the Central Universities (No.NJ2023032), the Scientific Research Starting Foundation of Nanjing University of Aeronautics and Astronautics (No.YQR21022), and the High Performance Computing Platform of Nanjing University of Aeronautics and Astronautics.

%======================================================== Acknowledgements ================================================================

% \section*{Acknowledgements}

\bibliography{custom}

%=========================================================== Appendix ======================================================================
\appendix

% \section{Appendix}
% \label{sec:appendix}

\section{Prompts}
\label{Appendix:prompt}
\begin{figure*}[htbp]
\center{
\includegraphics[width=0.8\linewidth]{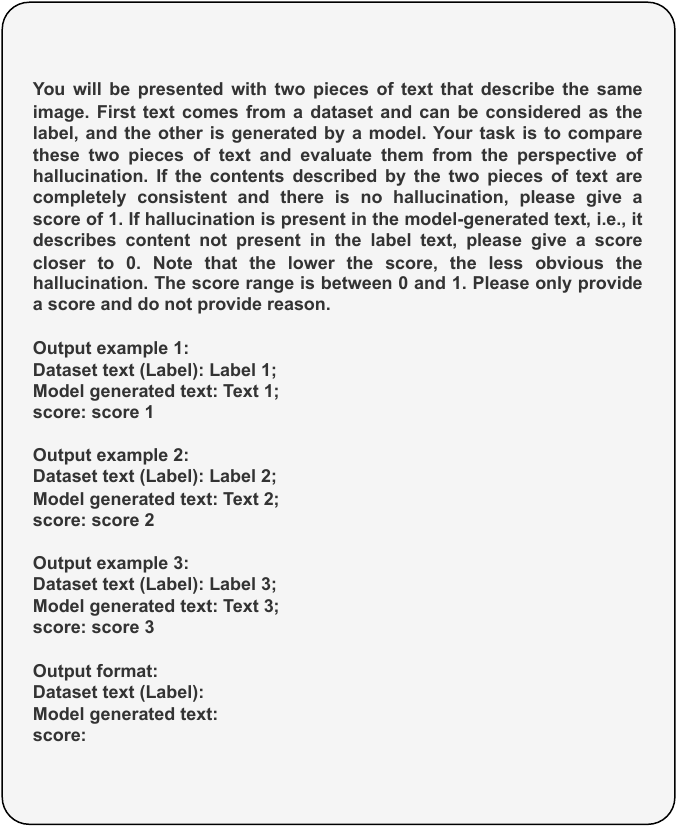}
}
\caption{\textbf{Prompt for generating sentence-level feedback score.}} 
\label{prompt_1}
\end{figure*}

\begin{figure*}[htbp]
\center{
\includegraphics[width=0.8\linewidth]{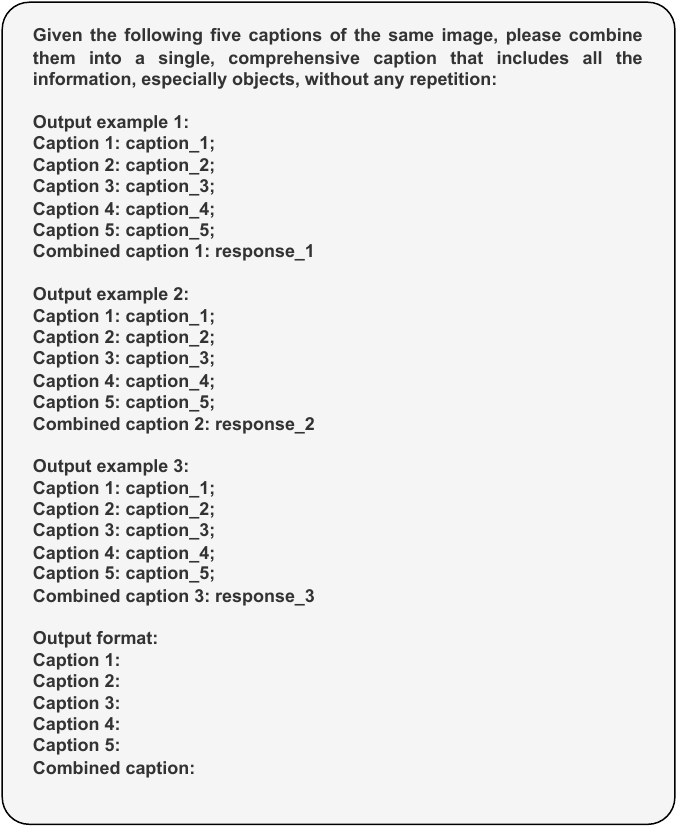}
}
\caption{\textbf{Prompt for synthesizing longer captions for each image based on corresponding five short captions.}} 
\label{prompt_2}
\end{figure*}

We show the prompt for generating sentence-level feedback score in Figure~\ref{prompt_1} and synthesizing longer captions for each image based on corresponding five short captions in Figure~\ref{prompt_2} with GPT-4.

\section{More Experimental Results}
\label{Appendix:results}
\subsection{More Experimental Results Compared to Existing Methods.}
For POPE benchmark in Table~\ref{tab:more_pop}, compared to existing methods (\citealp{liu2023mitigating, DBLP:journals/corr/abs-2403-14003, DBLP:journals/corr/abs-2403-14401}), our approach performs well on both the random and popular test sets, but falls short in terms of accuracy on the adversarial test set. This is because the method by \citet{zhou2023analyzing} uses multiple similar images for comparison, enabling multi-angle judgment of the test problem. In contrast, our proposed method requires less computational resources and is able to surpass the performance in terms of F1 score. While for CHAIR in Table~\ref{tab:more_chair}, compared to existing work, we are able to demonstrate better performance.

\subsection{Ablation Results About the Timing of Incorporating Hierarchical Feedback Learning.}
In Section~\ref{Further_analysis}, to investigate at which stage of training the integration of Hierarchical Feedback Learning can better enhance the model's anti-hallucination capabilities, we conduct an ablation study on the hyperparameter $c$. We show the additional results in Table~\ref{tab:more_timing}.

\subsection{Further Analysis About the Quality of the Generated Text.}
Considering that interventions in the decoding process can impact the quality of the generated text, we also conduct experiments to measure the impact caused by the vision-enhanced penalty decoding. Evaluation was carried out on the generated text from CHAIR, using the BLEU (\citealp{bleu}), ROUGE-L (\citealp{rouge}), METEOR (\citealp{METEOR}), and SPICE (\citealp{SPICE}) metrics, with the experimental results presented in Table~\ref{tab:nlg_eval}. It can be observed that our proposed vision-enhanced penalty decoding performs comparably to beam search in terms of text generation metrics, without demonstrating excessive decline. It even surpasses beam search on BLEU1, BLEU4, and ROUGE-L. As can also be seen from Table~\ref{tab:chair}, compared to beam search and opera decoding, our proposed method is able to maintain the length of the generated sentences as well. This elucidates that our method can maintain the quality of the generated text while mitigating hallucinations. Additionally, to verify whether HELPD can mitigate hallucinations while preserving general capabilities, we conduct evaluations on VQA-v2 (\citealp{DBLP:conf/cvpr/GoyalKSBP17}) and MME (\citealp{DBLP:journals/corr/abs-2306-13394}). As shown in Table~\ref{tab:VQA-v2} and Table~\ref{tab:MME-BENCH}, LVLMs with HELPD can maintain relatively stable performance across various metrics, demonstrating that the framework does not significantly impair the model's foundational abilities.

\section{Cases}
\label{Appendix:examples}
\begin{figure*}[t]
\center{
\includegraphics[width=1\linewidth]{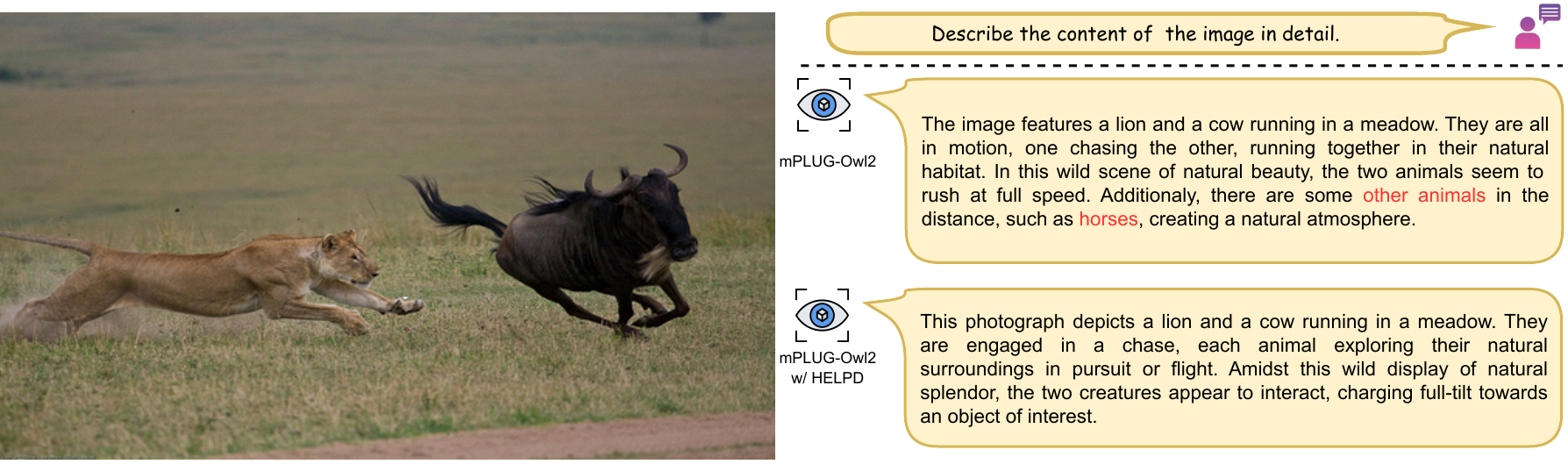}
}
\caption{\textbf{An illustrative case is presented to compare the output of mPLUG-Owl2 and mPLUG-Owl2 with HELPD.}} 
\label{case_2}
\end{figure*}

\begin{figure*}[t]
\center{
\includegraphics[width=1\linewidth]{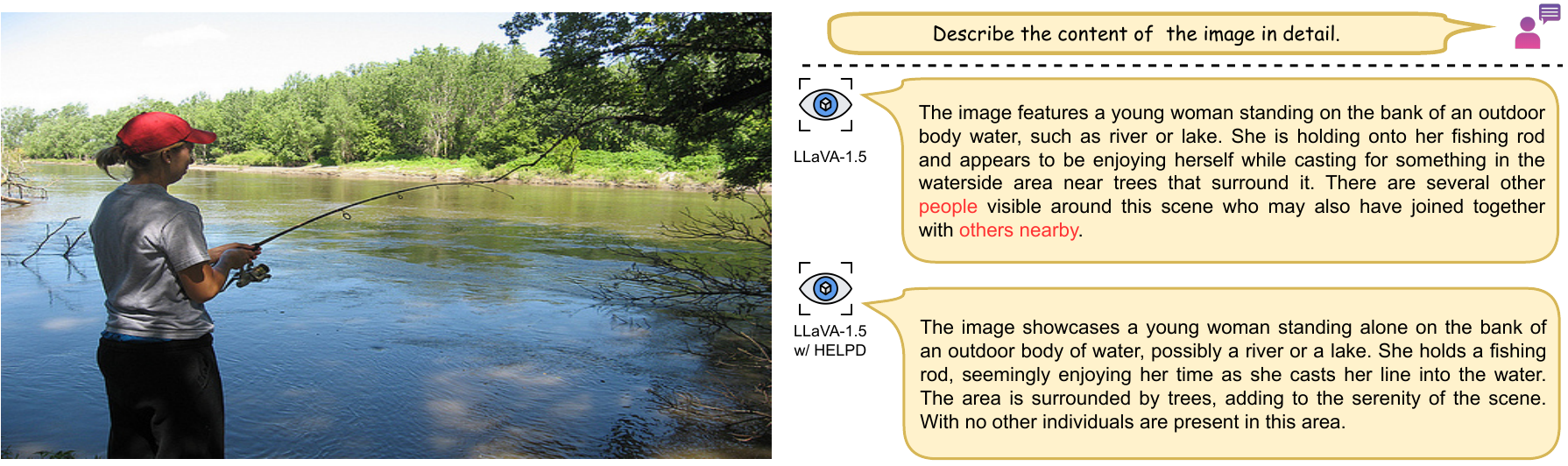}
}
\caption{\textbf{An illustrative case is presented to compare the output of LLaVA-v1.5 and LLaVA-v1.5 with HELPD.}} 
\label{case_3}
\end{figure*}

\begin{figure*}[t]
\center{
\includegraphics[width=1\linewidth]{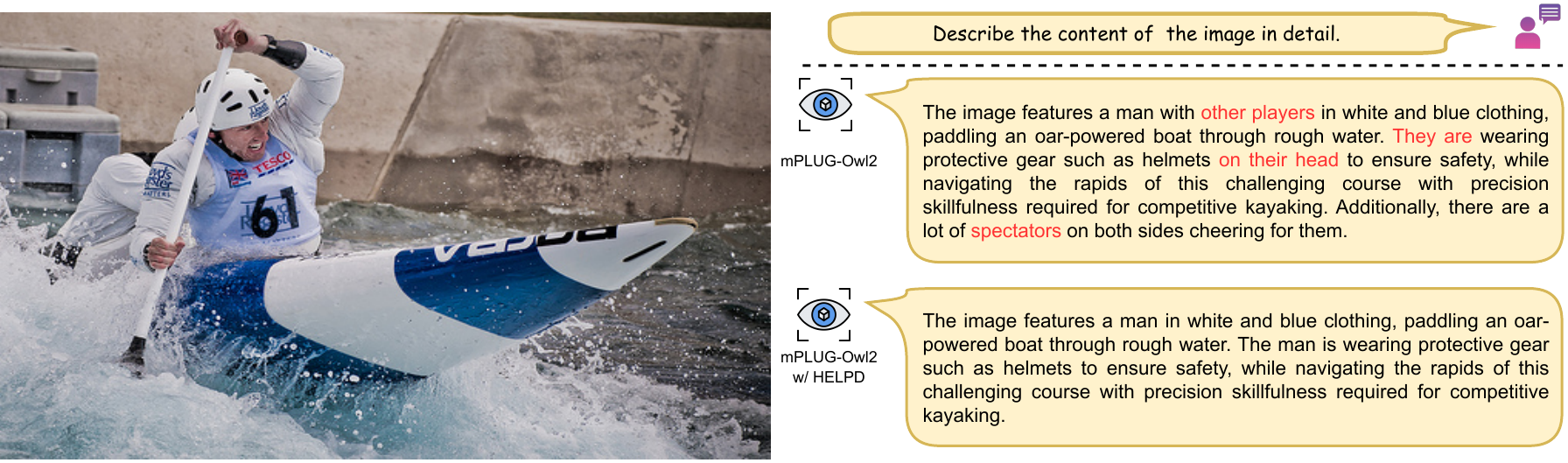}
}
\caption{\textbf{An illustrative case is presented to compare the output of mPLUG-Owl2 and mPLUG-Owl2 with HELPD.}} 
\label{case_4}
\end{figure*}

We show some generation cases in Figure~\ref{case_2}, \ref{case_3}, and \ref{case_4}.

\begin{table*}[t] 
\centering
\scalebox{1}{
\begin{tabular}{ll|c|cc|c|c}
\toprule
\textbf{POPE} & \textbf{Model}  &\textbf{$c$} & Precision & Recall & F1 Score & Yes (\%) \\
\midrule
\multirow{8}{*}{\textit{Random}} & \multirow{4}{*}{mPLUG-Owl2} & 0.6 & 88.7 & 86.1 & 87.3 & 46.5  \\
& & 0.7 & 88.9 & \textbf{85.9} & 87.3 & 46.4 \\
& & 0.8 & 89.9 & 85.6 & \textbf{87.7} & 47.6 \\  
& & 0.9 & \textbf{90.1} & 85.2 & 87.6 & 47.3\\ 
\cline{2-7}
&\multirow{4}{*}{LLaVA-1.5} & 0.6   & 87.0 & 92.1 & 89.5 & 52.9 \\
& & 0.7   & \textbf{87.8} & 91.9 & \textbf{89.9} & 52.4 \\
& & 0.8   & 86.2 & 93.1 & 89.5 & 54.6 \\  
& & 0.9   & 86.1 & \textbf{93.3} & 89.7 & 54.3\\

\midrule
\multirow{8}{*}{\textit{Popular}} & \multirow{4}{*}{mPLUG-Owl2} & 0.6 & 82.8 & 85.9 & 84.3 & 52.0  \\
& & 0.7 & 84.2 & \textbf{86.1} & 85.1 & 52.0 \\
& & 0.8 & 85.6 & 85.6 & \textbf{85.6} & 50.0\\    
& & 0.9 & \textbf{85.8} & 85.2 & 85.5 & 49.6\\ 
\cline{2-7}
&\multirow{4}{*}{LLaVA-1.5} & 0.6 & 80.1 & 93.3 & 86.2 & 58.2  \\
& & 0.7 & \textbf{81.4} & 92.0 & \textbf{86.4} & 56.5 \\
& & 0.8 & 79.7 & 93.0 & 85.8 & 58.3\\  
& & 0.9 & 79.5 & \textbf{93.5} & 85.9 & 58.8\\ 

\midrule
\multirow{8}{*}{\textit{Adversarial}} & \multirow{4}{*}{mPLUG-Owl2} & 0.6 & 79.2 & 85.9 & 82.5 & 54.1  \\
& & 0.7 & 79.8 & 85.9 & 82.7 & 54.2 \\
& & 0.8 & 80.2 & 85.6 & 82.8 & 53.3 \\  
& & 0.9 & 80.7 & 85.2 & 82.9 & 52.8 \\ 
\cline{2-7}
&\multirow{4}{*}{LLaVA-1.5} & 0.6 & 71.2 & 93.3 & 80.8 & 65.6  \\
& & 0.7 & \textbf{72.1} & 92.0 & \textbf{80.9} & 63.8 \\
& & 0.8 & \textbf{72.1} & 91.9 & 80.8 & 63.8\\  
& & 0.9 & 70.8 & \textbf{93.5} & 80.6 & 66.0\\ 

\bottomrule
\end{tabular}
}
\caption{\textbf{Additional ablation results on the timing of incorporating HELPD.}}
\label{tab:more_timing}
\end{table*}

\begin{table*}[t]
\centering
\scalebox{1}{
\begin{tabular}{ll|cccccc}
\toprule
\textbf{Model} & \textbf{Method} & B1 & B4 & M & R-L & S\\
\midrule
\multirow{3}{*}{\textit{LLaVA-1.5}}
& greedy            & 20.8 & 5.1 & 19.9 & 22.4 & 22.3\\
& beam$_\text{5}$   & 21.6 & 5.3 & \textbf{21.2} & 23.2 & \textbf{23.4}\\
& vep               & \textbf{22.3} & \textbf{5.5} & 20.8 & \textbf{23.3} & 22.9\\
\midrule

\multirow{3}{*}{\textit{mPLUG-Owl2}}
& greedy            & 18.9 & 4.8 & 16.7 & 17.1 & 16.9\\
& beam$_\text{5}$   & 19.8 & 5.1 & \textbf{17.2} & \textbf{19.7} & \textbf{18.3} \\
& vep               & \textbf{20.1} &\textbf{5.2} & \textbf{17.2} & 18.9 & 18.1 \\
\bottomrule
\end{tabular}
}
\caption{\textbf{Evaluation of the quality of the generated text from CHAIR.} B1, B4, M, R-L, and S are abbreviations for BLEU1, BLEU4, METEOR, ROUGE-L, and SPICE. ``vep'' represents the vision-enhanced penalty decoding.}
\label{tab:nlg_eval}
\end{table*}

\begin{table*}[t]
\centering
\scalebox{1}{
\begin{tabular}{l|cc|cc|cc}
\toprule
\multirow{2}{*}{\textbf{Method}} & \multicolumn{2}{c|}{\textit{Random}} & \multicolumn{2}{c|}{\textit{Popular}} & \multicolumn{2}{c}{\textit{Adversarial}} \\
\cline{2-7}
\multicolumn{1}{c|}{}                         & Accuracy         & F1 Score         & Accuracy          & F1 Score         & Accuracy            & F1 Score           \\
\midrule
\citet{liu2023mitigating}                                            & 87.3             & 87.3             & 73.4              & 80.1             & 65.0                & 73.9               \\
% \midrule
\citet{DBLP:journals/corr/abs-2403-14003}                                           & 81.2             & 65.6             & 73.9              & 67.3             & 68.2                & 75.4               \\
% \midrule
\citet{DBLP:journals/corr/abs-2403-14401}                                            & 87.5             & 86.1             & 85.1              & 84.8             & \textbf{81.7}       & 80.7               \\
% \midrule
HELPD (ours)                                        & \textbf{89.6}    & \textbf{89.8}    & \textbf{85.7}     & \textbf{86.6}    & 78.1                & \textbf{81.1}   \\
\bottomrule
\end{tabular}
}
\caption{\textbf{More experimental results on POPE benchmark.}}
\label{tab:more_pop}
\end{table*}

\begin{table*}[t]
\centering
\scalebox{1.1}{
\begin{tabular}{l|c|c|c}
\toprule
\textbf{Method} & CHAIR$_s$ $\downarrow$ & CHAIR$_i$ $\downarrow$ & Len\\
\midrule
\citet{zhou2023analyzing}               & 27.1  & 6.4 & 58.8 \\
\citet{DBLP:conf/emnlp/PoelCM22}               & 16.2  & 6.7 & 59.6 \\
\citet{DBLP:conf/acl/LiHFLEHZL23}               & 14.8  & 6.3 & 60.7 \\
\citet{liu2023mitigating}               & 13.8  & 5.9 & 60.1 \\
HELPD (ours)    & \textbf{9.6}   & \textbf{4.9} & \textbf{60.8} \\
\bottomrule
\end{tabular}
}
\caption{\textbf{More experimental results on CHAIR benchmark.}}
\label{tab:more_chair}
\end{table*}

\begin{table*}[t]
\centering
\scalebox{1}{
\begin{tabular}{l|c|c|c|c}
\toprule
\textbf{Method} & \textbf{Yes/No} &	\textbf{Number} &	\textbf{Other} &	\textbf{Overall}\\
\midrule
LLaVA-1.5 &	92.23 &	60.01 &	71.07 &	78.53 \\
LLaVA-1.5 (w/ ours) &	92.88 &	60.78 &	68.86 &	77.50 \\
mPLUG-Owl2 &	91.96 &	63.24	 & 70.51 &	79.05 \\
mPLUG-Owl2 (w/ ours) &	92.56 &	60.21 &	69.82 &	78.20 \\
\bottomrule
\end{tabular}
}
\caption{\textbf{Experimental results on VQA-v2 benchmark.}}
\label{tab:VQA-v2}
\end{table*}

\begin{table*}[t]
\centering
\scalebox{1.1}{
\begin{tabular}{l|c|c|c}
\toprule
\multirow{2}{*}{\textbf{Method}} & \multicolumn{2}{|c|}{\textbf{Category}} & \multirow{2}{*}{\textbf{Total}} \\
\cmidrule(){2-3}
\multicolumn{1}{c|}{}   & Perception    & Cognition & \multicolumn{1}{|c}{}       \\
\midrule
LLaVA-1.5 &	1508.5 &	355.7	 &	1864.2  \\
LLaVA-1.5 (w/ ours) &	1505.6 &	342.8 &	1848.4  \\
mPLUG-Owl2 &	1446.5 &	313.3	 & 1759.8  \\
mPLUG-Owl2 (w/ ours) &	1434.5 &	312.7 &	1747.2  \\
\bottomrule
\end{tabular}
}
\caption{\textbf{Experimental results on MME benchmark.}}
\label{tab:MME-BENCH}
\end{table*}

% \begin{figure*}[t]
% \center{
% \includegraphics[width=1\linewidth]{files/case_1.pdf}
% }
% \caption{\textbf{An illustrative case is presented to compare the output of LLaVA-v1.5 and LLaVA-v1.5 with HELPD.}} 
% \label{case_1}
% \end{figure*}

% \begin{table*}
%   \centering
%   \begin{tabular}{lll}
%     \hline
%     \textbf{Output}           & \textbf{natbib command} & \textbf{ACL only command} \\
%     \hline
%     \citep{Gusfield:97}       & \verb|\citep|           &                           \\
%     \citealp{Gusfield:97}     & \verb|\citealp|         &                           \\
%     \citet{Gusfield:97}       & \verb|\citet|           &                           \\
%     \citeyearpar{Gusfield:97} & \verb|\citeyearpar|     &                           \\
%     \citeposs{Gusfield:97}    &                         & \verb|\citeposs|          \\
%     \hline
%   \end{tabular}
%   \caption{\label{citation-guide}
%     Citation commands supported by the style file.
%     The style is based on the natbib package and supports all natbib citation commands.
%     It also supports commands defined in previous ACL style files for compatibility.
%   }
% \end{table*}

\end{document}